\documentclass[11pt]{article}
\usepackage{fullpage}
\usepackage{latexsym}
\usepackage{url}
\usepackage{subcaption}
\usepackage{natbib}
\usepackage{xcolor}

\usepackage{bm,amsmath,amssymb,amsthm}
\usepackage{graphicx}
\usepackage{hyperref}
\usepackage{amsfonts}
\usepackage{booktabs}
\usepackage{multirow}
\usepackage{Definitions}
\usepackage{subcaption}
\usepackage[ruled,vlined,linesnumbered]{algorithm2e}
\usepackage{color}
\usepackage{enumitem}

\usepackage{authblk}

\definecolor{mydarkblue}{rgb}{0,0.08,0.45}
\hypersetup{
    colorlinks,
    citecolor=mydarkblue,
    filecolor=mydarkblue,
    linkcolor=mydarkblue,
    urlcolor=mydarkblue
}

\graphicspath{{./figs/}}

\title{\textbf{$\text{KG}^2$: Learning to Reason Science Exam Questions \\ with Contextual Knowledge Graph Embeddings}}

\date{}

\author[1$^*$]{Yuyu Zhang}
\author[1\thanks{Both authors contributed equally to the paper.}]{Hanjun Dai}
\author[2]{Kamil Toraman}
\author[1]{Le Song}
\affil[1]{College of Computing, Georgia Institute of Technology}
\affil[2]{Korea Advanced Institute of Science and Technology}
\affil[1]{\texttt{\{yuyu.zhang, hanjun.dai, lsong\}@cc.gatech.edu}}
\affil[2]{\texttt{kvtoraman@kaist.ac.kr}}

\begin{document}

\maketitle

\begin{abstract}

The \textit{AI2 Reasoning Challenge (ARC)}, a new benchmark dataset for question answering (QA) has been recently released. ARC only contains natural science questions authored for human exams, which are hard to answer and require advanced logic reasoning. On the ARC Challenge Set, existing state-of-the-art QA systems fail to significantly outperform random baseline, reflecting the difficult nature of this task. In this paper, we propose a novel framework for answering science exam questions, which mimics human solving process in an open-book exam. To address the reasoning challenge, we construct contextual knowledge graphs respectively for the question itself and supporting sentences. Our model learns to reason with neural embeddings of both knowledge graphs. Experiments on the ARC Challenge Set show that our model outperforms the previous state-of-the-art QA systems.

\end{abstract}

\section{Introduction} \label{sec:intro}

Question answering (QA) has been a long-standing challenge in the field of artificial intelligence. Numerous research works have pushed forward techniques for building QA systems. Many existing approaches achieve high performance on benchmark datasets. However, most of the questions in those datasets only require surface-level reasoning, and do not reveal the full-scale complexity and challenge of the question answering problem.
% This may mislead the community into thinking that existing QA systems are good enough and we have almost solved QA, which is totally wrong. 
Recently, the \textit{AI2 Reasoning Challenge (ARC)} has been proposed~\citep{clark2018think}, which is designed to pose a challenge to the QA community. On the ARC Challenge Set, several state-of-the-art QA systems, including leading neural models from the well-known SQuAD and SNLI tasks, only perform slightly better than the random baseline. This striking observation has demonstrated that QA is still far from being solved.

Why it is so difficult to answer the questions in the ARC Challenge Set? 1) ARC consists of natural science questions, namely questions authored for human exams. All of these questions are drawn from real exams; 2) In order to encourage progress on hard questions, a Challenge Set has been partitioned from ARC. To be more specific, if a question could not be correctly answered by neither an information retrieval (IR) method nor a word co-occurrence method, it is sorted into the Challenge Set, otherwise the Easy Set. To illustrate the difference, consider the following two examples from both sets respectively, where the bold answers correspond to the correct choices:  

\begin{itemize} %[noitemsep,nolistsep,topsep=0pt,leftmargin=*]
  \item {\bf ARC Easy Set:} 
  \noindent \begin{quote}
  \textit{Which property of air does a barometer measure? (A) speed \textbf{(B) pressure} (C) humidity (D) temperature}
  \end{quote}
  
  This question is correctly answered by both the IR and word co-occurrence methods.\footnote{Note that even it is correctly answered by only one of them, ARC would exclude it from the Challenge Set.} The IR method finds sentences relevant to the correct answer in the reference corpus, e.g., ``\textit{Air pressure will be measured with a barometer}''. Due to the substantial word overlap, the question can be easily solved. Similarly, the word co-occurrence method finds that ``barometer'' and ``pressure'' co-occur frequently in the corpus, leading to the correct answer.  
  
  \item {\bf ARC Challenge Set:} 
  \noindent \begin{quote}
  \textit{Which property of a mineral can be determined just by looking at it? \textbf{(A) luster} (B) mass (C) weight (D) hardness}
  \end{quote}
  
Neither the IR method nor the word co-occurrence method can correctly answer this question. There are no sentences in the corpus similar to ``\textit{A material's luster can be determined by looking at it}''. Also, ``mineral'' often co-occurs with distractor options (e.g., mass, hardness), which confuses the word co-occurrence method.
\end{itemize}

From the examples above, we see that surface-level reasoning methods are not able to solve questions in the Challenge Set, even the required knowledge is already covered in the reference corpus. The ARC Corpus, a large science-related text corpus collected from the Web and released together with ARC, mentions knowledge relevant to about 95\% of the ARC Challenge questions~\citep{clark2018think}. However, the IR method with the ARC Corpus, as listed in Table~\ref{table:score}, only achieves 20.26 test score, which underperforms the random baseline. Collecting more sentences into the corpus would not solve the challenge. Actually we tried to use the entire Web as the reference corpus with Google Search API, and select the answer option with the most number of hits. This only slightly improves the score to 21.58.

To tackle the ARC Challenge, we believe that there is no shortcut to get around advanced logic reasoning and deeper text comprehension. These questions target at students of age 8 through 13 years old, and should be relatively easy for human to solve. For an adult with basic reasoning capability, even she forgets about the knowledge learned in grade school, she can still ace most of these questions in an open-book exam, by searching relevant supporting texts and reasoning over them.

Inspired by the human problem solving process, we propose a neural reasoning engine named $\text{KG}^2$ for answering science exam questions: read the question, generate hypothesis by combining the question stem and answer option, find supporting sentences in the corpus, and verify the hypothesis. For effective and efficient reasoning, we represent both hypothesis and supporting sentences in knowledge graphs. For example, in the supporting graph, ``\textit{luster}'' is linked to ``\textit{brightness}'', and ``\textit{brightness}'' connects to ``\textit{look}'', which is consistent with the hypothesis graph. Therefore, such reasoning patterns on graphs can be learned by our differentiable neural engine.
Experiments on the ARC Challenge Set show that our model achieves score that surpasses the previous state-of-the-art results.

%\Le{Add a more refined description over the reasoning process including the knowledge graph reasoning. For instance, use the ARC challenge set example you had above.} 

%\Le{Design your differentiable neural programming engine according to the reasoning pattern in the above reasoning example. You only described the knowledge graph part.} 

%\Le{Get the end-to-end version working as well.}
%\Yuyu{I find that we achieve the best performance when considering all tuples generated by open IE. Any filtration will actually hurt the performance. I think what we consider as noise might have helpful information for the reasoning.}

%\Le{Need a more refined description of your experimental results instead of just a single number. Is it learning any reasoning? What pattern do you learn? What kind of question did you solve? What is still difficult?} 

In summary, the contributions of this work are:
1) We propose a novel differential neural programming framework for reasoning about science exam questions;
2) Our method sets the new state of the art on the ARC Challenge Set;
3) We decompose the remaining difficulties towards solving the ARC Challenge, facilitating the community to engage with the dataset and progress on the challenging task.

\section{Related Work}

\noindent \textbf{Science QA:} For elementary science QA, simple IR-based methods have been proposed for science exams~\citep{clark2016combining}. Markov Logic Networks~\citep{richardson2006markov} has been used to reason over a small set of logical rules~\citep{khot2015markov}. \citet{jansen2016s} has analyzed knowledge and inference requirements for science exam questions.

The work most related to us is DGEM~\citep{khot2018scitail}, a neural entailment model which also employs Open IE to generate hypothesis graph. Our key contributions over DGEM: 1) DGEM is designed for single sentence entailment, while we aggregate multiple supporting sentences for reasoning; 2) DGEM has no structured representation of supporting facts, while our model learns to reason over the paired hypothesis and supporting graphs together.

%\Le{How is it different from those on memory network? How is it different from Lao Ni's neural programming engine? or neural module network?} 
%\Yuyu{Lao Ni's paper is very different: they learn to parse the question into an executable program on an existing knowledge graph, but they don't use any open-domain text as supports.}

\noindent \textbf{Graph Embedding:} We employ graph embedding techniques for reasoning over knowledge graphs. Graph embedding has provided the representational flexibility for neural models in many NLP tasks, such as dialog system~\citep{he2017learning}, question answering~\citep{zhang2017variational}, link prediction~\citep{bordes2013translating} and triple classification~\citep{feng2016gake}. In our paper, we extend this technique to mimic the reasoning process on graph ranking problem. 

\section{Task} \label{sec:task}

The ARC Challenge Set consists of science exam questions $\mathcal{D}=\big\{ q_i, \  \big(c_i^{(1)}, \ldots, c_i^{(m)}\big), \ a_i \big\}_{i=1}^n$, where $q_i$ is the question stem, $c_i^{(j)}$ is the $j$-th answer option corresponding to $q_i$ (typically 4-way multiple choices), and $a_i$ is the label of correct answer. Both $q_i$ and $c_i^{(j)}$ are in text format. Among the multiple choices, only one of them is the correct answer and others are distractors. With the question stem and options, the goal is to find the correct answer. Accompanied with ARC, the ARC Corpus is also provided, providing 14M science-related sentences from the Web with knowledge relevant to ARC. The use of the ARC Corpus is optional for the ARC Challenge.

\section{Approach} \label{sec:approach}

%The human solving process of science exam questions motivates us to decompose the QA system into multiple stages. Our framework consists of four components: 1) Generating hypothesis from the question stem and each answer option (Section~\ref{subsec:gen_hypo}); 2) Searching potential supports from the corpus (Section~\ref{subsec:search}); 3) Constructing knowledge graphs for both the hypothesis and supports (Section~\ref{subsec:kg}); 4) Learning to reason with knowledge graph embeddings (Section~\ref{subsec:embedding}).

%\subsection{Human Reasoning Process} 
%
%\Le{Provide a more formal description of human reasoning process} 
%
%\Le{Formalize the reasoning process into a procedure. For the part with uncertainty or with complex input information, neuralize the operation.}
%
%\Yuyu{Seems no enough space for this section in a 4-page short paper. We can add it in full paper.}

\subsection{Generating Hypothesis} \label{subsec:gen_hypo}

A hypothesis $h$ is a statement that combines a question stem $q$ and an answer option $c$, which helps us understand what is being asked and what is the target to be verified. For example, consider the question stem ``\textit{Which of these occurs due to the rotation of Earth?}'' and one of the answer options ``\textit{day and night}''. The hypothesis to be generated from them should be: ``\textit{Day and night occurs due to the rotation of Earth}''.

%\Le{example good.} 

%\Le{Besides word description, can it be formalized with very simple notation or symbols. This may not be on emnlp style but need to be done this way to make precise.}

To automatically generate hypothesis, we first identify the wh-word (e.g., which, what, where, etc.) in the question stem, and replace it with the answer option. If there is no wh-word found, we just append the answer option behind the question stem. We create several rules to handle special cases and make hypothesis more natural. For example, ``\textit{Which of these}'' and ``\textit{Which of the following}'' should be replaced as a whole when they appear in the question stem. We successfully generate hypothesis for most questions, however, there are still a few corner cases requiring advanced rewording, which should be negligible.

%\Le{isn't it you also generate graph from this? Can you use neural program to help potential corner case?}
%\Yuyu{The corner cases are very rare, and special cases, such as ``when'' is not a question word but the beginning of a clause. Also, we don't have enough supervision to train neural program here.}

\subsection{Searching Potential Supports} \label{subsec:search}

To verify a hypothesis, we look for supports in the reference corpus. Although the corpus is typically gigantic, we only need to focus on a tiny part of it, which is relevant to the question we are solving. Therefore, we use the generated hypothesis as a query to search the entire corpus. The top retrieved sentences are treated as potential supports for the hypothesis. In order to efficiently search the corpus, we build a local search engine on top of ElasticSearch~\citep{gormley2015elasticsearch}. Since the corpus sentences are not as clean as questions, we filter noisy sentences that contain negation words (e.g., not, except, etc.) or unexpected characters or simply too long, and then pick up the top 20 sentences for verifying the hypothesis.
 
%\Le{The search process can be made more general by have indexing or embedding for query $q_i$ and the content $v_j$. You pre-designed or learned a similarity function $s(q_i, v_j)$. One thing you can do is to have a ElasticSearch match score as baseline line score and then a neuralized score as the increment. That is $s(q_i, v_j) = s_1(q_i, v_j) + s_2(q_i, v_j| \theta)$. In some sense, this is like residual network, which there is a neural network which augments a deterministic operation. Similar ideas can also be used in earlier replacement process and generate query graph.}

%\Yuyu{Good point. This is kind of a neural search engine to adjust the ranking scores. The worry here is that what if we expand our corpus, and the model has to be retrained end-to-end? I think it makes sense to over-retrieve sentences by IR, and then re-rank them by neural models before we pick up the top few ones.}

\subsection{Constructing Knowledge Graphs} \label{subsec:kg}

Many questions in the ARC Challenge Set require advanced reasoning on multiple supporting sentences. To aggregate knowledge across sentences, we employ Open IE~\citep{banko2007open,christensen2011analysis,pal2016demonyms} v4~\footnote{\url{https://github.com/allenai/openie-standalone}} to extract relation triples from each sentence, and collect them to construct a contextual knowledge graph.

More specifically, each relation triple is represented as $T(s, p, o_i)$, where $s$ is the subject, $p$ is the predicate, and $o_i$ is the $i$-th object. We construct the graph by adding nodes $s$, $p$ and $o_i$, and adding directed edges with labels $subj$ and $obj$. If there is adverbial of time or location extracted by Open IE, we add an edge with label $time$ or $loc$ in the knowledge graph. Words in each graph node are lemmatized. Similarly, we construct another knowledge graph for the corresponding hypothesis, which is paired with the supporting knowledge graph. Refer to Appendix~\ref{apdx:graph} for examples of our generated graphs.

%\Le{I still don't get why can't one build the knowledge graph for the supporting documents first and then one can index subgraph for the search process. For the query sentence, the graph needs to be build on the fly. The basic idea of how OpenIE work and the key formalization needs to be explained and this will help us neural the operation using residual network type of idea.}

%\Yuyu{We can do that, but it's not flexible if we want to use external knowledge or even the entire Web for document retrieval. I will read original papers of Open IE and we may improve that by neural model.}

\subsection{Learning with Graph Embeddings} \label{subsec:embedding}

%\Le{Why can't you modularize the reasoning process, and use Markov logic network as inspiration which need to hand design feature. We can learn feature for the inference step. I'd like to see this analogy made clear.} 

Given a question $q$ and a candidate choice $c$, we construct the corresponding hypothesis graph $G^{hypo}_{q, c}$ and supporting graph $G_{q, c}^{supp}$ by aggregating the relation triples mentioned in Section~\ref{subsec:kg}. Thus, choosing the right answer for question $q_i$ becomes a graph ranking problem. A good graph scoring function $f: G^{hypo} \times G^{supp} \mapsto \mathbb{R}$ should assign the highest score to the correct hypothesis-supporting graph pair. Without loss of generality, we use point-wise ranking objective, where $f(\cdot)$ becomes a binary classifier.  

%\begin{eqnarray}
% & \max_{f: \mapsto [0, 1]}\frac{1}{nm} \sum_{i=1}^n \sum_{j=1}^m  \\
%	& \II(c_i^{(j)} = a_i) \log \big(f(G^{hypo}_{q_i, c_i^{(j)}}, G^{supp}_{q_i, c_i^{(j)}}) \big) + \\
%	& \Big(1 - \II(c_i^{(j)} = a_i)\Big) \log \big(1 - f(G^{hypo}_{q_i, c_i^{(j)}}, G^{supp}_{q_i, c_i^{(j)}}) \big)
%\end{eqnarray}
To implement the graph scoring function, we adapt the recent advances in graph embedding~\citep{dai2016discriminative, gilmer2017neural} to our problem. Specifically, let $G = (V, E)$ be a knowledge graph, and $V_p \in V$ be the set of predicate nodes. We associate each node $v \in V$ with an embedding vector $\mu_v$ that captures the local information, which is computed recursively using the equation:
\begin{equation}
	\mu_v^{(t)} = h\big( \xb_v, \mu_v^{(t-1)}, \{(\mu_u^{(t-1)}, e_{u, v})\}_{ (u, v, e_{u, v}) \in E}\big)
	\label{eq:graph_embed}
\end{equation} 
Here $\xb_v$ encodes the text feature of node generated by LSTM that is jointly trained with the supervision. The edge type $e_{u, v}$ can be \emph{time}, \emph{loc}, etc. We use a two-layer neural network for the function $h(\cdot)$. Eq.~\eq{eq:graph_embed} iterates for $T$ steps, and we use $\mu_v = \mu_v^{(T)}$ as the node embedding representation.
Finally, the scoring function $f(\cdot)$ is defined as:
\begin{eqnarray}
	f(G^{hypo}, G^{supp}) = & f(\{ \mu_u\}_{u \in V_p^{hypo}}, \{ \mu_v\}_{v \in V_p^{supp}}) \nonumber \\
	= & \sigma\Big( \max_{u, v} \frac{\mu_u^\top \mu_v}{\|\mu_u\| \|\mu_v\|}  - 0.5 \Big),
	\label{eq:score}
\end{eqnarray}
where $\sigma(\cdot)$ is the sigmoid function, and the $-0.5$ shift is used to center the matching score at zero. Eq.~\eq{eq:score} is making max inner product search between all pairs of predicate node embeddings. This mimics the procedure of reasoning on the most relevant hypothesis and corresponding supporting evidence, since each embedding vector already captures the information within its $T$-hop neighborhood.

\section{Experiments} \label{sec:exp}
%\Le{The error decomposition is good. They should tell us how to improve our current method. It seems that we need to consider getting more information in, and learning complex reasoning better or from somewhere else, and without relying solely on openIE. This will be good direction to formalize the problem for the summer.} 

%\Yuyu{That's right. Retrieval bias is also a problem, which should be improved by constructing smarter queries and probably searching iteratively to refine the supporting sentences. For Open IE, we may improve that by dependency parser.}

We compare our method against several recently published baseline models, including state-of-the-art neural models from the well-known SQuAD and SNLI tasks.

\subsection{Setup}

We use the ARC Challenge Set~\citep{clark2018think} for all experiments. This dataset consists of 2,590 questions drawn from a variety of human exams. We use the original train / development / test split. The test set is held-out for model evaluation, which contains 1,172 questions. For each question, a QA system receives one point if it selects the correct answer, and $1/k$ points if it reports a $k$-way tie (i.e., chooses multiple answers) that includes the correct answer. The ARC Corpus can be optionally used for all models.

\subsection{Baselines}

\noindent \textbf{Guess-all / Random:} This naive baseline just selects all answer options, getting $1/k$ scores for each question with $k$ answer options. Random selecting will also converge to this score after enough trials.

\noindent \textbf{IR-ARC:} IR-based method sends question stem plus each option as a query to a search engine. For IR-ARC, the search engine is built on top of the ARC Corpus, and the search score is determined by the ElasticSearch score of the top retrieved sentence. The option with the highest search score is finally selected.

\noindent \textbf{IR-Google:} This is similar to IR-ARC, but uses Google Search API~\footnote{\url{https://developers.google.com/custom-search}} to retrieve documents from the entire Web, instead of just searching on the ARC Corpus. IR-Google uses the number of hits as the search score.

\noindent \textbf{TableILP:} This method~\citep{khashabi2016question} performs table-based reasoning, which is formulated as an Integer Linear Program (ILP).

\noindent \textbf{TupleInference:} This model~\citep{khot2017answering} searches for graph that best connects the terms in the question with an answer choice via the knowledge extracted by Open IE.

\noindent \textbf{DecompAttn:} It is a neural entailment model~\citep{parikh2016decomposable} adapted to multiple-choice QA by assigning entailment score to the pair of hypothesis and single supporting sentence~\citep{clark2018think}. The answer option with the highest score is selected. DecompAttn is a top performer on SNLI~\citep{bowman2015large}.

\noindent \textbf{DGEM-OpenIE:} DGEM~\citep{khot2018scitail} is also a neural model for sentence-level entailment, but uses Open IE to create structured representation of the hypothesis. On the SciTail task~\citep{khot2018scitail}, DGEM is a top performer. In \cite{clark2018think}, there is another version of DGEM, which uses a proprietary parser together with Open IE and achieves 27.11 test score. For fair comparison, we only list publicly available models in Table~\ref{table:score}.

\noindent \textbf{BiDAF:} This model~\citep{seo2016bidirectional} is for span prediction QA, and has been adapted to multiple-choice QA~\citep{clark2018think}. BiDAF is a top performer on SQuAD~\citep{rajpurkar2016squad}.

\subsection{Results and Analysis}

\begin{table}[t]
\setlength\tabcolsep{12pt}
\centering
\caption{Test performance of different QA systems on the ARC Challenge Set. The ARC Corpus is used in DecompAttn, DGEM, BiDAF and $\text{KG}^2$.}
\label{table:score}
\begin{tabular}{@{}lc@{}}
\toprule
\textbf{Method} & \textbf{Test Scores} \\ \midrule
IR-ARC & 20.26 \\
IR-Google & 21.58 \\
TupleInference & 23.83 \\
DecompAttn & 24.34 \\
Guess-all / Random & 25.02 \\
DGEM-OpenIE & 26.41 \\
BiDAF & 26.54 \\
TableILP & 26.97 \\ \midrule
$\text{KG}^2$ & \textbf{31.70} \\ \bottomrule
\end{tabular}
\end{table}

Table~\ref{table:score} summarizes the test scores of all baseline models and our method. It is striking to see that none of the baseline methods perform significantly better than the random baseline, where the 95\% confidence interval is $\pm2.5\%$. Our method $\text{KG}^2$ achieves 31.70, which substantially improves the previous state of the art by 17.5\%.

\begin{figure}[t]
\centering
\includegraphics[width=0.53\textwidth]{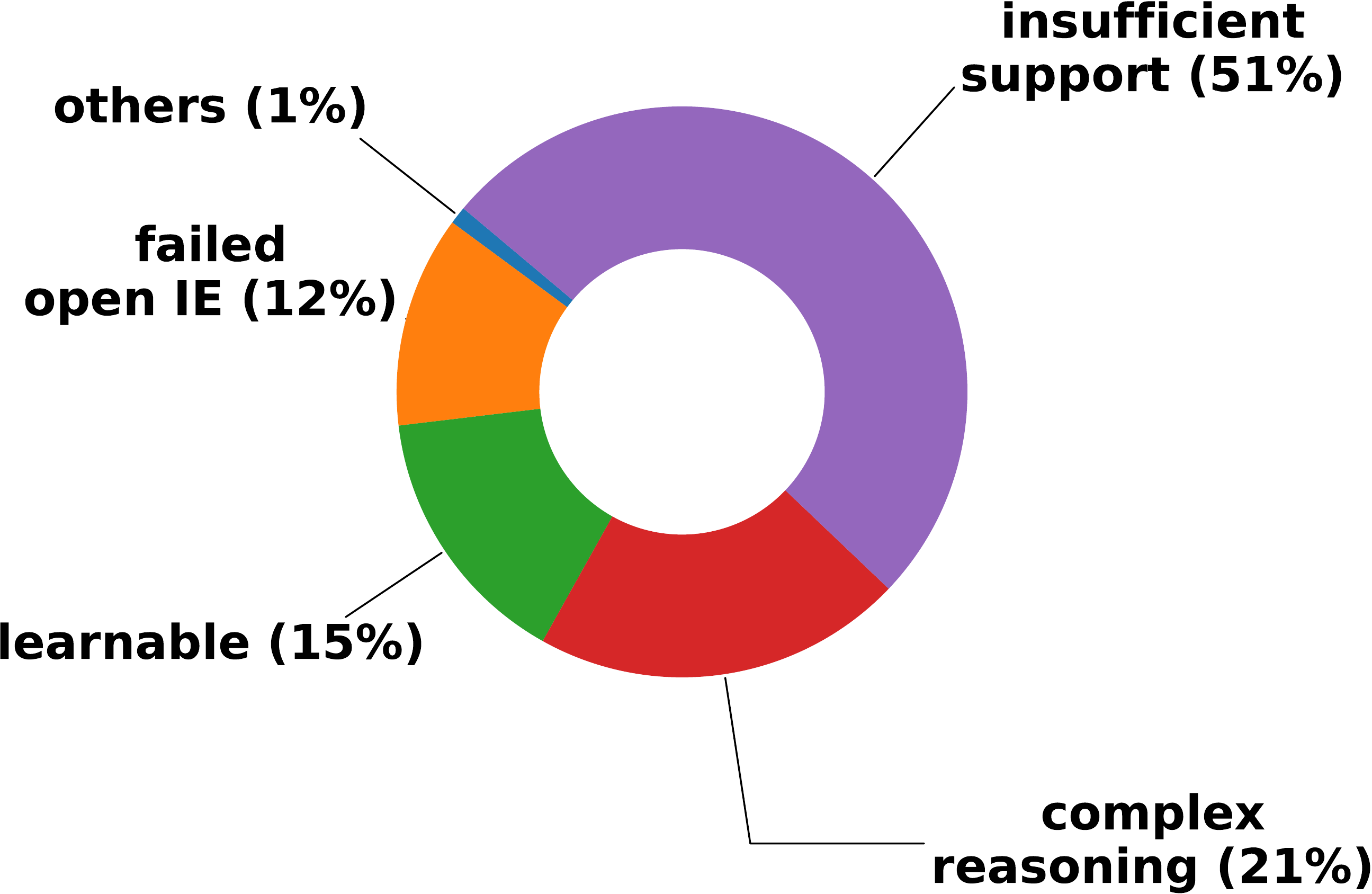}
\caption{Distribution of various difficulties in solving the ARC Challenge Set.}
\label{fig:donut}
\end{figure}

Nevertheless, we are still far from ``passing'' the exam. To dissect the difficulties, we randomly sample 100 questions for investigation and report the results in Figure~\ref{fig:donut}. More than half of the questions are lack of support: even human couldn't solve them by only referring to the supporting sentences. This may be caused by the limited coverage of the corpus, and the retrieval bias where sentences with low word overlap can partially explain concept which is indispensable for reasoning. External knowledge sources may help on these questions. 12\% questions have lost key information in graph, due to the failure of Open IE. Sentence parsing may be helpful since it reserves more text. 21\% questions require very complex reasoning, and only 15\% questions are ``learnable'' given the current framework. This gives us an estimated upper bound when we correctly answer all the learnable questions, and just randomly guess the others, which should be 36.25. Improving 
the learning algorithm should bring our current result closer to this upper bound.

\section{Conclusion and Future Work}

We present a neural reasoning engine for answering science exam questions, which learns to reason over contextual knowledge graphs. Experimental results show that our method outperforms existing QA systems on the ARC Challenge Set. In the future, we will explore how to exploit external knowledge sources, and try to improve the quality of open IE by sentence parsing.

\clearpage

% include your own bib file like this:
\bibliography{qa}
\bibliographystyle{plainnat}

\clearpage

\appendix

\noindent {\LARGE\bf Appendix}

\section{Examples of Knowledge Graphs} \label{apdx:graph}

To illustrate how we construct knowledge graphs from hypothesis and supporting sentences, here we present some examples.

\begin{figure}[h]
\centering
\begin{subfigure}{0.48\textwidth}
	\centering
	\includegraphics[width=0.4\textwidth]{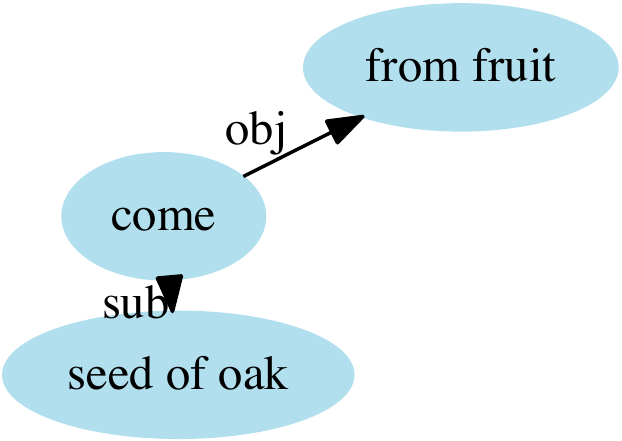}
	\caption{Knowledge graph for hypothesis}
	\label{fig:kg_hypo}
\end{subfigure}
\begin{subfigure}{0.48\textwidth}
	\centering
	\includegraphics[width=\textwidth]{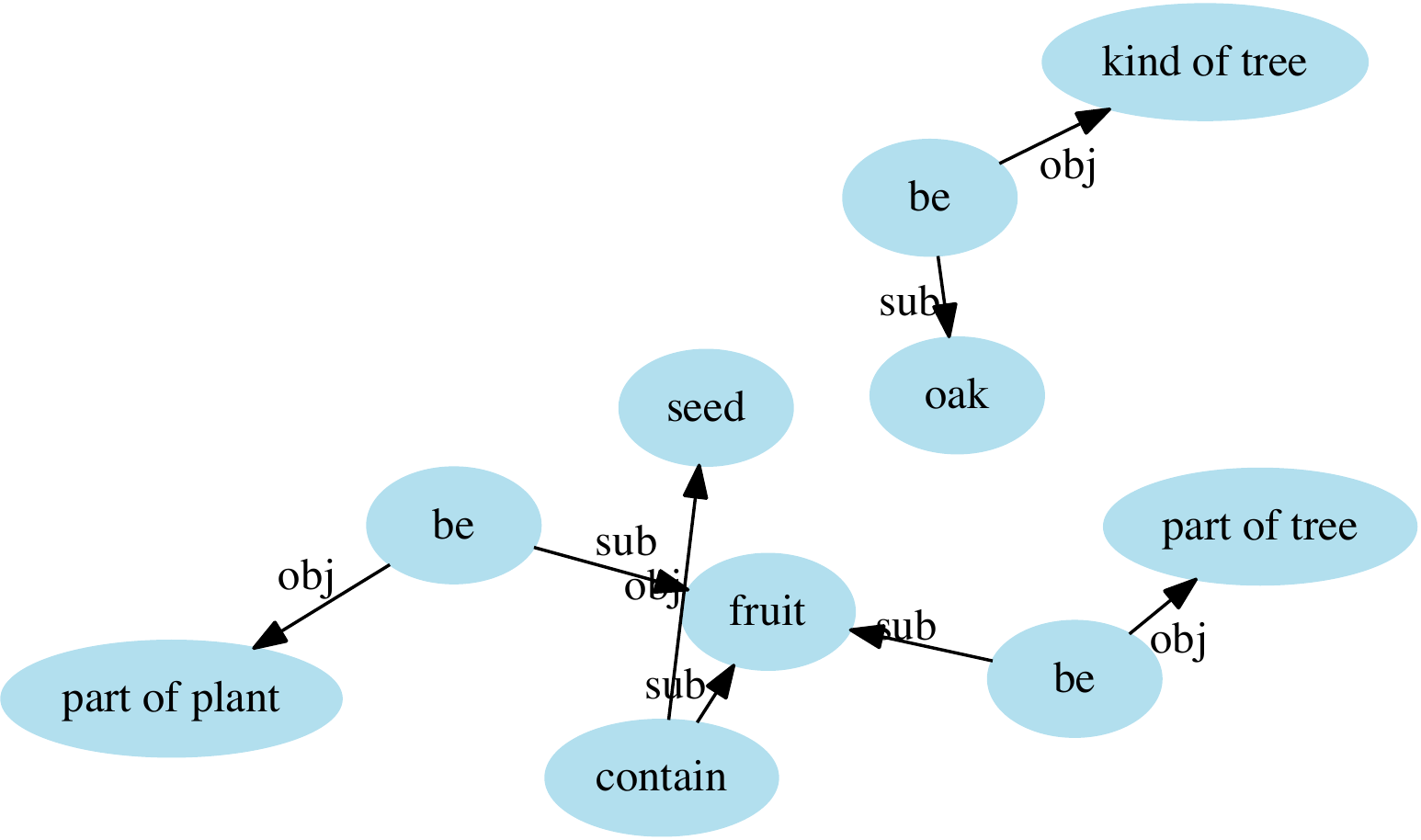}
	\caption{Knowledge graph for supports}
	\label{fig:kg_support}
\end{subfigure}
\caption{Example of knowledge graphs for paired hypothesis and supports.}
\label{fig:kg_example}
\end{figure}

We first show a relatively simple example in Figure~\ref{fig:kg_example}. We see a pair of hypothesis and supporting graphs. The hypothesis is ``\textit{seed of oak comes from fruit}'', as shown in Figure~\ref{fig:kg_hypo}. Note that the verb ``\textit{comes}'' is lemmatized and becomes ``\textit{come}'' in the graph. The supporting knowledge graph is plotted in Figure~\ref{fig:kg_support}, where we obtain knowledge including ``\textit{fruit contains seed}'', ``\textit{fruit is part of tree}'', and ``\textit{oak is kind of tree}''. With the supporting knowledge graph, we should be able to infer that the hypothesis is true.

\begin{figure*}[h]
\centering
\begin{subfigure}{\textwidth}
	\centering
	\includegraphics[width=0.26\textwidth]{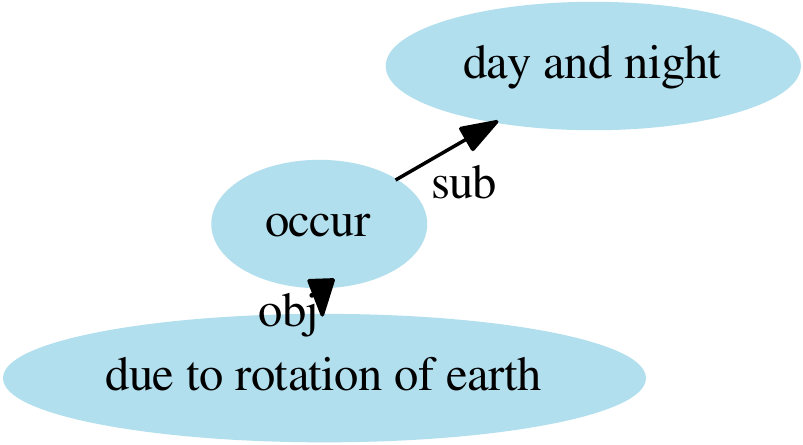}
	\caption{Knowledge graph for hypothesis}
	\label{fig:kg_hypo2}
\vspace{5mm}
\end{subfigure}
\begin{subfigure}{\textwidth}
	\centering
	\includegraphics[width=\textwidth]{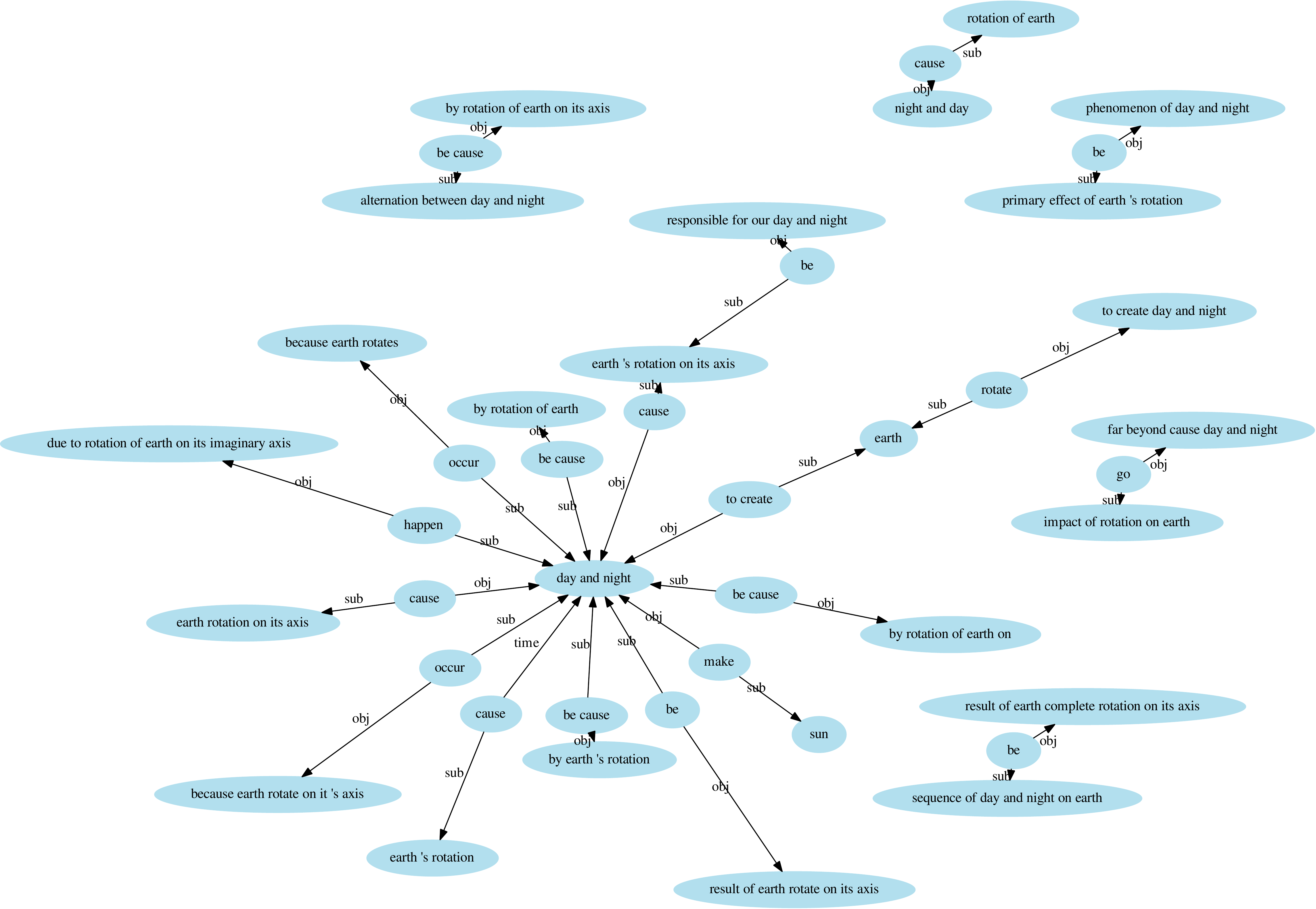}
	\caption{Knowledge graph for supports}
	\label{fig:kg_support2}
\end{subfigure}
\caption{Another example of knowledge graphs for paired hypothesis and supports.}
\label{fig:kg_example2}
\end{figure*}

Note that the knowledge graphs can be very complicated when the question stem has multiple sentences, or there is rich information in the supporting sentences extracted by Open IE. We show another example in Figure~\ref{fig:kg_example2}, which has a heavier supporting graph than the previous example. This is actually common in the ARC Challenge Set. In this example, the hypothesis is ``\textit{day and night occurs due to rotation of earth}'', as plotted in Figure~\ref{fig:kg_hypo2}. Looking at the supporting graph in Figure~\ref{fig:kg_support2}, we can find key information for this question, such as ``\textit{day and night occurs because earth rotates}'', ``\textit{day and night causes earth rotation on its axis}'', ``\textit{day and night is caused by earth's rotation}'', etc. With the supporting knowledge graph, we should have necessary information to verify the hypothesis.

\end{document}